\documentclass[10pt]{article} 

\usepackage[preprint]{rlj} 

\usepackage{amssymb}            
\usepackage{mathtools}          
\usepackage{mathrsfs}           
\usepackage{graphicx}           
\usepackage{subcaption}         
\usepackage[space]{grffile}     
\usepackage{url}                
\usepackage{lipsum}             
\usepackage{threeparttable}     
\usepackage{booktabs}

\title{Sample-Efficient Ensemble Reinforcement \\ Learning at Scale}

\setrunningtitle{Sample-Efficient ERL at Scale}

\author{Yoav Feigenbaum, Dmitrii Krylov, and Roy Fox}

\emails{yfeigenb@uci.edu, dkrylov@uci.edu, royf@uci.edu}

\affiliations{
\textbf{Department of Computer Science, University of California, Irvine}\\

}


\begin{document}

\maketitle  

\begin{abstract}
  Ensemble methods such as MeanQ have been shown to improve the sample efficiency of relatively small model-free deep reinforcement learning (RL) architectures, including DQN and similarly sized variants such as Rainbow. However, to our knowledge, the effects of ensembling on large-scale architectures such as Bigger Better Faster (BBF) have not been studied. In this work, we present \textbf{MeanQ-BBF}, a hybrid model-free algorithm that combines the principles of MeanQ and BBF. Through statistically rigorous evaluation on the Atari 100K benchmark, we show that in contrast to naive scaling of the network size of BBF, which degrades performance, MeanQ-BBF is able to leverage the increase in model capacity to improve performance over BBF. Moreover, MeanQ-BBF does this in a computationally efficient way by organizing the added parameters into largely independent learners. Finally, we discuss the performance-cost tradeoff in the context of practical sample-efficient RL.
\end{abstract}


\section{Introduction}

Reinforcement Learning (RL) is a powerful framework for sequential decision-making that has many useful applications in domains such as robotics \citep{kober2013}, healthcare \citep{yu2021}, and autonomous vehicles \citep{aradi2022}. Its remarkable success in complex game environments, including Go \citep{silver2016}, Atari \citep{mnih2013}, and StarCraft II \citep{vinyals2019}, has largely been driven by deep neural networks that approximate value functions and policies \citep{mnih2015,schulman2017}. However, deep RL approaches demand significant amounts of environment interaction to achieve effective training, posing a significant barrier in settings where data collection is costly, risky, or time-intensive. As a result, improving sample efficiency is a critical goal for RL algorithms intended to be deployed in real-world environments.

Prior work has explored two promising directions for improving sample efficiency. \citet{liang2022} introduced MeanQ, which uses an ensemble of networks to reduce variance in value estimates by combining their predictions. \citet{schwarzer2023} introduced BBF, which scales the model architecture through wider and deeper neural networks to improve representational power and learning stability.

In this work, we propose MeanQ-BBF, a hybrid RL algorithm that combines variance reduction via ensembling with a large-scale deep RL architecture, a combination that, to our knowledge, has not been studied in the context of sample efficiency. Although these two methods are largely orthogonal, subtle interactions between them influence their combined effectiveness, which we analyze and address through careful design choices. We focus on the model-free setting, which forgoes explicit dynamics modeling in favor of a simpler and more computationally efficient learning pipeline, which we discuss in the context of recent model-based approaches in Section~\ref{sec:related}.

We demonstrate that MeanQ-BBF improves upon the sample efficiency of BBF, as evaluated on the Atari 100K benchmark. Moreover, we show that naively scaling the size of the networks used for BBF is not sufficient on its own, and argue that the ensemble structure is necessary for preserving sample efficiency when scaling. Finally, we discuss the broader implications of applying ensemble RL to large-scale architectures, considering both the performance gains and the associated computational costs.


\section{Background}

Reinforcement learning (RL) formalizes sequential decision-making as a Markov
Decision Process (MDP) \citep{sutton1998}, defined by the tuple
$(\mathcal{S}, \mathcal{A}, P, R, \gamma)$, where $\mathcal{S}$ is the state
space, $\mathcal{A}$ the action space, $P$ the transition dynamics, $R$ the
reward function, and $\gamma \in [0,1)$ a discount factor controlling the
relative weighting of immediate versus future rewards. At each timestep $t$,
an agent observes state $s_t$, selects action $a_t$ according to a policy
$\pi$, receives scalar reward $r_t$, and transitions to successor state
$s_{t+1}$. The agent's goal is to find a policy $\pi^*$ maximizing the
expected discounted return $G_t = \mathbb{E}_\pi [ \sum_{k=0}^{\infty} \gamma^k r_{t+k+1} ]$.

A central quantity in RL is the action-value function $Q^\pi(s, a)$,
representing the expected discounted return from taking action $a$ in state
$s$ and following policy $\pi$ thereafter. The optimal action-value function
$Q^*(s, a)$ satisfies the Bellman optimality equation:
\begin{equation}
    Q^*(s, a) = \mathbb{E}\!\left[
        r + \gamma \max_{a'} Q^*(s', a') \;\middle|\; s, a
    \right],
\end{equation}
from which the optimal policy is recovered greedily as
$\pi^*(s) = \arg\max_a Q^*(s, a)$. During training, action selection
typically follows an $\varepsilon$-greedy policy, taking a random action with
probability $\varepsilon$ and the greedy action otherwise, balancing
exploration with exploitation.

In high-dimensional environments such as Atari, the Q-function is approximated
by a neural network $Q(s, a;\,\theta)$, giving rise to Deep Q-Networks (DQN)
\citep{mnih2015}. The network is trained by minimizing the mean squared
temporal difference (TD) error between the current Q-value estimates and
bootstrap targets computed using a periodically updated frozen copy of the
network, called the target network. Past transitions are stored in a replay
buffer $\mathcal{D}$ and sampled during training, decoupling the update
distribution from the agent's current trajectory. The ratio of gradient
updates to environment steps, known as the replay ratio, controls how
intensively the agent learns from collected data; DQN uses a replay ratio
of 0.25, performing one gradient update every four environment steps.
Rather than bootstrapping from a single-step TD target, $n$-step returns
can be used to accumulate rewards over $n$ steps before bootstrapping,
a quantity referred to as the update horizon, which reduces bias at
the cost of increased variance.

A known pathology of this formulation is overestimation bias, caused by using
the same network for both selecting and evaluating the greedy action. Double
DQN \citep{hasselt2016} corrects this by decoupling the two roles: the online
network $\theta$ selects the greedy action, while the target network $\theta^-$
evaluates it:
\begin{equation}
    y = r + \gamma\, Q\!\left(s',\, \arg\max_{a'} Q(s', a';\, \theta);\, \theta^-\right).
\end{equation}

\section{Related Work}
\label{sec:related}

\subsection{Variance Reduction with MeanQ}

Sample-efficient RL agents must extract accurate value estimates from noisy, limited data conditions, making variance reduction a particularly useful property in this setting. MeanQ addresses this through an ensembling approach: state-action value estimates, used both to select the maximizing action and to compute training targets, are averaged across Q-networks, thereby reducing estimation variance without introducing substantial architectural complexity \citep{liang2022}. All of the MeanQ ensemble members share a single replay buffer, but each samples from it using its own priorities, which reflect that member's own Bellman errors. As a result, the members train on different minibatches, which preserves some decorrelation between their updates despite the shared data. 

On the Atari 100K benchmark, MeanQ was shown to outperform comparably sized baselines, including DQN \citep{mnih2015}, Rainbow DQN \citep{hessel2018}, SUNRISE \citep{lee2021}. However, this evaluation was confined to relatively small neural network architectures, leaving open the question of whether ensemble-based variance reduction remains effective at larger scales.

\subsection{Architecture Scaling with BBF}

In a distinct line of research, Bigger, Better, Faster (BBF) demonstrated that scaling to significantly wider and deeper neural networks could achieve superhuman performance on the Atari 100K benchmark, far surpassing the smaller architectures on which MeanQ had been evaluated \citep{schwarzer2023}.

Crucially, BBF showed that naive architectural scaling alone is insufficient. Drawing on several independently developed techniques, effective sample-efficient performance also relies on hard resets of the convolutional layers for regularization \citep{doro2023}, an increased replay ratio \citep{fedus2020}, a dynamic update horizon \citep{kearns2000}, a higher discount factor \citep{françoislavet2016}, and weight decay through use of the AdamW optimizer \citep{loshchilov2019}. BBF also incorporates Self-Predictive Representations (SPR) \citep{schwarzer2021}, an auxiliary self-supervised objective that trains the agent to predict latent representations of future states without requiring additional labeled data. This encourages the agent to develop richer and more generalizable internal representations earlier in training, further improving data efficiency. Together, these modifications allowed BBF to realize the benefits of larger-scale architectures.

Despite the significant gains in sample efficiency made by BBF, the interaction between BBF-style architecture scaling techniques and ensemble methods such as MeanQ remains largely unexplored, particularly whether ensemble-based variance reduction can be used to compensate for the degradation in sample efficiency that arises when naively scaling model capacity.

\subsection{Model-Based Learning}
Beyond these model-free approaches, recent advances in model-based RL have demonstrated impressive performance and sample efficiency by leveraging learned world models to generate synthetic experience and facilitate planning, with methods such as DreamerV3 \citep{hafner2024} and EfficientZero V2 \citep{wang2024} achieving strong results on standard benchmarks. However, these benefits often come with additional computational and memory costs associated with model learning and planning, while inaccuracies in learned dynamics can limit the effectiveness of imagined experience \citep{wang-zirui2024, barkley2025}. These trade-offs motivate continued investigation into improving the sample efficiency of model-free RL, which avoids explicit dynamics modeling and planning and can offer a simpler and more computationally efficient learning pipeline, particularly when data is scarce.

\section{Method}
\label{sec:method}

\definecolor{inhcol}{HTML}{1a6b30}
\definecolor{exccol}{HTML}{111111}

\newcommand{\inh}{\textcolor{inhcol}{$\checkmark$}}
\newcommand{\absent}{{\color{exccol}$-$}}
\newcommand{\na}{{\color{exccol}\textit{N/A}}}

\begin{table}[ht]
  \centering
  \caption{Technique-level breakdown of MeanQ, BBF, and MeanQ-BBF\@.
            Status symbols are defined in the notes below.}
  \setlength{\tabcolsep}{14pt}
  \renewcommand{\arraystretch}{1.35}
  \begin{threeparttable}
    \label{tab:method-breakdown}
    \begin{tabular}{@{\hspace{6pt}} l ccc @{\hspace{6pt}}}
      \toprule
      \textbf{Technique}
        & \textbf{MeanQ}
        & \textbf{BBF}
        & \textbf{MeanQ-BBF} \\
      \midrule
      \quad Ensemble-mean targets       & \inh    & \absent & \inh \\
      \quad Shared exploration policy   & \inh    & \na     & \inh \\
      \quad Replay buffer               & \inh    & \inh    & \inh\tnote{$\dagger$} \\
      \quad Replay sampling priorities  & \inh\tnote{$\dagger$}    & \inh     & \inh\tnote{$\dagger$} \\
      \quad Lagging target network      & \absent & \inh    & \absent \\
      \quad Large networks              & \absent & \inh    & \inh \\
      \quad Hard resets                 & \absent & \inh    & \inh \\
      \quad High replay ratio           & \absent & \inh    & \inh \\
      \quad SPR auxiliary loss          & \absent & \inh    & \inh\tnote{$\dagger$} \\
      \quad Dynamic update horizon      & \absent & \inh    & \inh \\
      \quad Increased discount $\gamma$ & \absent & \inh    & \inh \\
      \quad Weight decay                & \absent & \inh    & \inh \\
      \quad Double DQN                  & \absent & \inh    & \absent \\
      \bottomrule
    \end{tabular}
    \begin{tablenotes}[flushleft]\footnotesize
      \item
        \inh~Present\enskip
        \absent~Not present\enskip
        \na~Concept does not apply
      \item[$\dagger$] Maintained independently for each ensemble member.
    \end{tablenotes}
  \end{threeparttable}
\end{table}

MeanQ-BBF is a hybrid algorithm that integrates the ensemble-based variance reduction of MeanQ into the scalable BBF framework to improve sample efficiency in large-scale deep RL. Since the ensembling strategy outlined by MeanQ is straightforward and modular compared with the numerous implementation techniques employed by BBF, we use BBF as the foundation for MeanQ-BBF and wrap it with the MeanQ algorithm. This approach leverages BBF's scalable architecture while incorporating the variance-reducing benefits of ensembling, and requires several targeted design decisions to ensure that the statistical independence of ensemble members is preserved and the variance reduction benefits of ensembling are fully realized.

\textbf{Ensemble Size.} We set the ensemble size to 5, consistent with the value used by MeanQ \citep{liang2022}, balancing the benefit of variance reduction against computational cost. We have not tuned this hyperparameter.

\textbf{Replay Buffer Design.} Following MeanQ, we maintain independent replay sampling priorities for each ensemble member to reduce potential correlations between their learning signals. Moreover, we use a separate replay buffer for each member to further decouple their learning trajectories. Despite the added memory cost, we prioritize independence with the aim of improving variance reduction.

\textbf{Self-Predictive Representations (SPR).} Following the same reasoning, we instantiate a separate SPR module \citep{schwarzer2021} for each ensemble member rather than sharing one across the ensemble, further preserving independence between members. 

\textbf{Double DQN.} We disable the Double DQN \citep{hasselt2016} component of BBF, since the ensemble mean already reduces overestimation bias by averaging Q-value estimates across members, making the bias-overcorrection mechanism of Double DQN more harmful here than helpful.

\textbf{Target Networks.} We also disable the use of lagging target networks, as MeanQ demonstrated that the ensemble mean provides sufficient stability in value estimates to substitute for the smoothing effect of a lagging target network \citep{liang2022}.

\section{Results}
\label{sec:results}

\subsection{Experimental Setup}

We evaluate MeanQ-BBF on the Atari 100K benchmark \citep{kaiser2020}. The benchmark limits each agent to 100,000 environment interaction steps, roughly equivalent to two hours of human gameplay, and evaluates performance across a selection of 26 Atari video games that do not require advanced exploration or agent memory. In this limited-data setting, the game score achieved by agents is a useful metric for sample efficiency. For statistical robustness, each experiment was run with five independent seeds per game.

While BBF achieves its best performance at a replay ratio of 8~\citep{schwarzer2023}, under computational resource constraints, we adopt a replay ratio of 2 across all experiments to ensure fair comparison between agents. The original BBF results indicate that performance at replay ratio 2 remains competitive, and since higher replay ratios consistently outperform lower ones, we expect MeanQ-BBF to similarly benefit from a higher replay ratio.

We leave all remaining hyperparameters unchanged from the published BBF agent, and reuse this hyperparameter configuration for our MeanQ-BBF sub-agents, with the exception of Double DQN disabled (see section \ref{sec:method}). Since the size of the ensemble is set to five, the MeanQ-BBF agent contains exactly five times the model parameters as an individual BBF. The exact parameter counts vary slightly depending on the size of the action space for a given game, but are approximately 39M for BBF and 195M for MeanQ-BBF.

To assess the effects of naive model scaling, we introduce a ``Scaled BBF'' variant, in which the network width scale is increased from 4 to 20 without any additional mitigation techniques, producing a structure identical to BBF but with a model size comparable to MeanQ-BBF (approximately 210M parameters). This design choice is motivated by prior BBF findings that scaling to large architectures without supplementary techniques, such as convolutional resets, increased replay ratios, or the default weight decay, leads to performance degradation~\citep{schwarzer2023}, and that further scaling beyond BBF size remains to be explored.

\subsection{Evaluation}

All scores are human-normalized such that a score of 0 corresponds to a random policy and a score of 1 corresponds to average human performance, to enable fair comparison and averaging across games with different scoring systems.

Aggregate performance is assessed using the interquartile mean (IQM) score. The IQM is calculated as the mean of the middle 50\% of human-normalized scores across all games and seeds, discarding the lowest and highest quartiles. Compared to the simple mean, IQM is more robust to outliers and skewed distributions, which is particularly important on the Atari 100K benchmark where certain games such as Breakout, Boxing, and Krull are relatively easy for RL agents to master, producing exceptionally high human-normalized scores that would otherwise disproportionately inflate the mean. Full per-game results are provided in Appendix \ref{sec:appendix1}.

\begin{figure*}[t!]
    \centering
    \begin{subfigure}[t]{0.5\textwidth}
        \centering
        \includegraphics[height=2.2in]{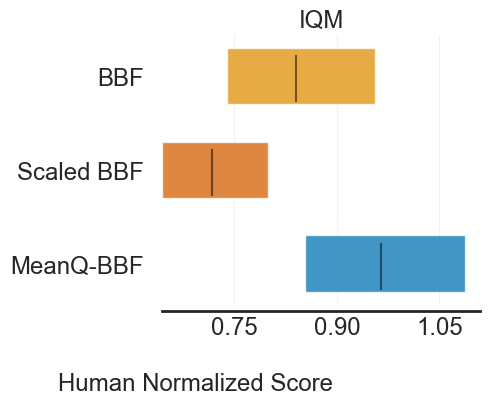}
    \end{subfigure}%
    ~ 
    \begin{subfigure}[t]{0.5\textwidth}
        \centering
        \includegraphics[height=2.2in]{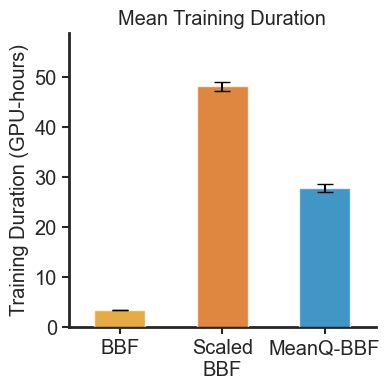}
    \end{subfigure}
    \caption{Aggregate human-normalized IQM score \textbf{(left)} and average training GPU-hours needed \textbf{(right)} for BBF, Scaled BBF, and MeanQ-BBF on the Atari 100K benchmark. 95\% confidence intervals are displayed as shaded regions around the IQM point estimate and as error bars for the training durations.}
    \label{fig:aggregates}
\end{figure*}

Results of our experiments provide IQM score estimates of 0.84 for BBF, 0.72 for Scaled BBF, and 0.96 for MeanQ-BBF. Around these point estimates, we construct 95\% confidence intervals using repeated bootstrapping (Agarwal et al., 2021), with the resulting intervals displayed in Figure \ref{fig:aggregates} as colored regions around each estimate.

Despite having the largest model capacity, Scaled BBF performs poorly, consistent with BBF's finding that naively scaling deep RL architectures without mitigation techniques such as well-tuned periodic resets and weight decay leads to performance degradation, likely due to overfitting \citep{schwarzer2023}. In sharp contrast, MeanQ-BBF outperforms BBF on 15/26 environments, and Scaled BBF on 19/26 with a 34\% higher aggregate IQM score than Scaled BBF.

\begin{figure*}[h]
    \center
     \includegraphics[width=0.6\textwidth]{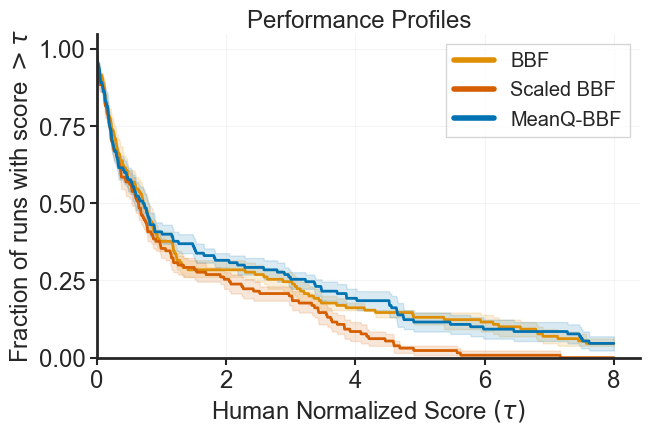}
    \caption{Performance profile comparing BBF, Scaled BBF, and MeanQ-BBF across all 26 Atari 100K games with 5 seeds each, following the \texttt{rliable} framework \citep{agarwal2021}.}
    \label{fig:perf_profile}
\end{figure*}

Figure~\ref{fig:perf_profile} displays a performance profile following the \texttt{rliable} framework \citep{agarwal2021}, plotting the fraction of runs across all 26 games and all seeds that surpassed each human-normalized score threshold, where a higher curve indicates better performance. Unlike summary statistics such as the mean or IQM, the performance profile captures the full distribution of results across the different game environments and provides a more robust basis for comparison. MeanQ-BBF achieves the highest performance profile curve overall, most notably in medium-difficulty tasks. It also overwhelmingly dominates Scaled BBF across the full score range, indicating that MeanQ-BBF attains better sample efficiency within the same interaction budget.

\subsection{Computational Cost}

Consistent with the original MeanQ algorithm~\citep{liang2022}, computing targets for an ensemble of size $K$ requires $O(K^2)$ target evaluations per step, since each member's target computation evaluates all $K$ members' value estimates at the next states in its minibatch. For $K=5$, this results in 25 target forward passes per step versus BBF's single pass, contributing to an overall computational cost of approximately $8.11\times$ that of BBF (Figure~\ref{fig:aggregates}). While this superlinear cost is inherent to the ensemble target computation, MeanQ-BBF remains significantly more efficient than Scaled BBF, which requires approximately $14.03\times$ the GPU-hours as BBF, consistent with the original MeanQ finding that the quadratic cost is manageable at small ensemble sizes.

\section{Conclusion}
\label{sec:conclusion}

In this work, we presented MeanQ-BBF, a hybrid algorithm combining the ensemble-based variance reduction of MeanQ with the large-scale architecture of BBF, and evaluated it on the Atari 100K benchmark. While naively scaling the network size of BBF degraded performance, our empirical results indicate that the ensemble structure provided by MeanQ-BBF not only prevents this performance degradation, but modestly outperforms BBF. Since this ensemble structure is largely complementary to the network scaling innovations of BBF, we contend that ensemble-based variance reduction is a viable technique for improving sample efficiency in large-scale, model-free deep RL architectures.

We also emphasize that these sample efficiency gains come at a non-negligible computational cost, which grows superlinearly with the ensemble size. However, MeanQ-BBF remains significantly more computationally efficient than a BBF agent naively scaled to a similar model size, making it a practical option in domains where environment interaction data is scarce.

The interaction between ensemble variance reduction and model-based RL remains an open question. Recent model-based methods have demonstrated strong sample efficiency through learned world models, and it is natural to ask whether ensembling provides complementary benefits in that setting as well. This work merits further investigation into ensemble methods as a general tool for scaling sample-efficient RL.

\subsubsection*{Acknowledgments}
\label{sec:ack}
The research work of author Roy Fox is supported by the National Science Foundation and the Binational Science Foundation

\bibliography{main}
\bibliographystyle{rlj}

\appendix

\section{Score Distributions for Atari 100K Games}
\label{sec:appendix1}

\begin{figure}[h]
    \centering
    \includegraphics[width=0.85\textwidth]{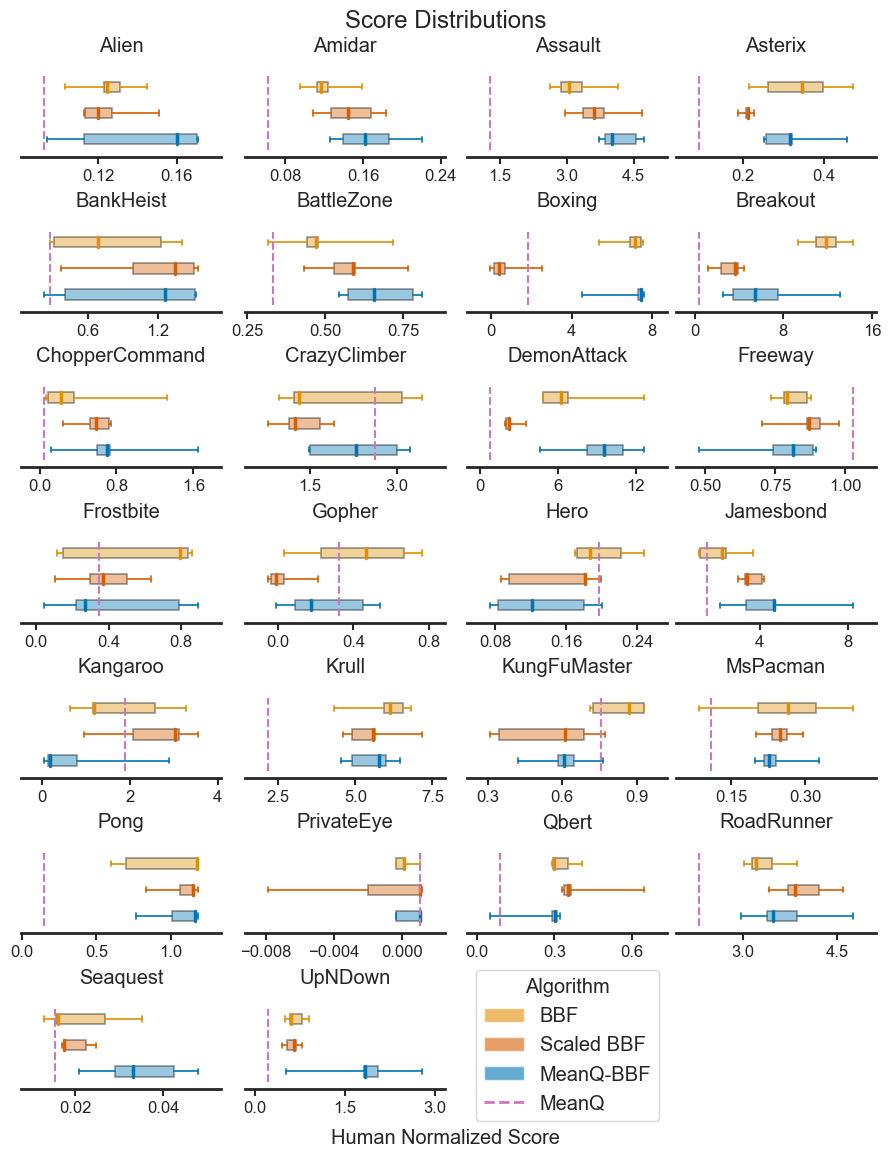}
    \caption{Box plots displaying the results for five seeds on each of the 26 Atari 100K games. MeanQ per-game mean scores as published are indicated by dashed lines}
    \label{fig:all_scores}
\end{figure}

\begin{table}[htbp]
    \centering
    \caption{Average raw scores for each algorithm on each game at 100K interaction steps. Standard deviations across five runs are shown in parentheses. Data for MeanQ is reported directly from \citet{liang2022}. The best performance for each environment is bolded.}
    \label{tab:raw_scores}
    \resizebox{\textwidth}{!}{%
        \begin{tabular}{l | l | l | l l | l l | l l | l l}
        \toprule
        \textbf{Environment} & \textbf{Random} & \textbf{Human} & \multicolumn{2}{c|}{\textbf{MeanQ}} & \multicolumn{2}{c|}{\textbf{BBF}} & \multicolumn{2}{c|}{\textbf{Scaled BBF}} & \multicolumn{2}{c}{\textbf{MeanQ-BBF}} \\
        \midrule
        Alien          & 184.8   & 7128.0  & 864.8            & (309.7)   & 1092.4           & (93.0)    & 1088.1           & (97.1)    & \textbf{1204.5}  & (221.0)   \\
        Amidar         & 11.8    & 1720.0  & 112.5            & (28.3)    & 214.1            & (35.6)    & 256.7            & (46.6)    & \textbf{291.9}   & (58.3)    \\
        Assault        & 233.7   & 742.0   & 894.4            & (120.7)   & 1891.5           & (270.9)   & 2141.5           & (297.9)   & \textbf{2393.7}  & (207.1)   \\
        Asterix        & 248.8   & 8503.0  & 971.5            & (246.7)   & \textbf{3023.0}  & (766.3)   & 1950.4           & (105.2)   & 2872.4           & (615.5)   \\
        BankHeist      & 15.0    & 753.0   & 215.6            & (282.4)   & 590.7            & (342.1)   & \textbf{864.0}   & (322.9)   & 743.9            & (413.8)   \\
        BattleZone     & 2895.0  & 37188.0 & 13980.0          & (3225.0)  & 19242.0          & (4506.0)  & 22630.0          & (3769.0)  & \textbf{25806.0} & (3733.1)  \\
        Boxing         & 0.3     & 12.0    & 22.2             & (9.2)     & \textbf{82.7}    & (9.5)     & 8.9              & (11.1)    & \textbf{82.7}             & (14.2)    \\
        Breakout       & 0.9     & 30.0    & 13.0             & (6.2)     & \textbf{341.6}   & (47.7)    & 91.6             & (33.9)    & 186.6            & (107.9)   \\
        ChopperCommand & 671.0   & 7388.0  & 1118.8           & (330.5)   & 3529.0           & (3077.1)  & 4526.2           & (1194.0)  & \textbf{5818.4}  & (3258.8)  \\
        CrazyClimber   & 7339.5  & 35829.0 & \textbf{76490.0} & (15610.2) & 60966.6          & (26073.1) & 44645.4          & (10033.4) & 68339.2          & (18240.6) \\
        DemonAttack    & 140.0   & 1971.0  & 1636.8           & (287.6)   & 13049.0          & (5238.8)  & 4541.4           & (1067.9)  & \textbf{16929.5} & (4964.3)  \\
        Freeway        & 0.0     & 30.0    & \textbf{30.4}    & (0.9)     & 24.0             & (1.6)     & 25.6             & (2.7)     & 22.6             & (4.5)     \\
        Frostbite      & 74.0    & 4334.7  & 1557.8           & (1214.6)  & \textbf{2418.9}  & (1468.4)  & 1693.8           & (766.6)   & 1955.4           & (1428.2)  \\
        Gopher         & 245.9   & 2412.0  & 956.0            & (207.5)   & \textbf{1189.0}  & (586.3)   & 318.8            & (205.5)   & 793.0            & (455.9)   \\
        Hero           & 224.6   & 30826.0 & 6893.2           & (1192.8)  & \textbf{6972.9}  & (911.7)   & 5453.1           & (1418.2)  & 4950.0           & (1506.2)  \\
        Jamesbond      & 29.2    & 303.0   & 466.2            & (80.1)    & 629.8            & (248.5)   & 1016.1           & (123.2)   & \textbf{1292.6}  & (556.5)   \\
        Kangaroo       & 42.0    & 3035.0  & 5714.0           & (4012.9)  & 5343.8           & (2952.4)  & \textbf{7644.4}  & (2762.7)  & 2496.8           & (3205.2)  \\
        Krull          & 1543.3  & 2666.0  & 3913.4           & (185.7)   & \textbf{7952.5}  & (937.7)   & 7544.7           & (954.3)   & 7497.1           & (757.4)   \\
        KungFuMaster   & 616.5   & 22736.0 & 17232.0          & (8019.1)  & \textbf{18982.4} & (2185.5)  & 12501.8          & (4163.9)  & 13838.6          & (2484.2)  \\
        MsPacman       & 235.2   & 6952.0  & 1032.0           & (164.5)   & \textbf{1998.5}  & (705.4)   & 1958.7           & (209.2)   & 1918.7           & (301.5)   \\
        Pong           & -20.4   & 15.0    & -15.4            & (4.0)     & 13.4             & (9.2)     & \textbf{17.2}    & (4.5)     & 16.6             & (5.6)     \\
        PrivateEye     & 26.6    & 69571.0 & \textbf{100.0}   & (0.0)     & 33.0             & (36.5)    & -65.2            & (243.6)   & 60.0             & (49.0)    \\
        Qbert          & 166.1   & 13455.0 & 1352.2           & (571.9)   & 4540.1           & (607.4)   & \textbf{5555.6}  & (1615.1)  & 3555.8           & (1351.5)  \\
        RoadRunner     & 0.0     & 7845.0  & 18012.5          & (7229.0)  & 26165.8          & (2317.1)  & \textbf{30968.8} & (3195.1)  & 28899.4          & (4729.6)  \\
        Seaquest       & 61.1    & 42055.0 & 721.8            & (62.4)    & 964.8            & (352.1)   & 901.8            & (131.0)   & \textbf{1523.8}  & (400.9)   \\
        UpNDown        & 488.4   & 11693.0 & 2982.5           & (560.9)   & 8037.6           & (1664.2)  & 7401.9           & (1296.7)  & \textbf{20607.3} & (8151.1)  \\
        \bottomrule
        \end{tabular}%
    }
\end{table}

\end{document}